\documentclass[preprint,12pt]{elsarticle}

\usepackage{graphicx}
\usepackage{amssymb}
\usepackage{siunitx}
\usepackage{subfig}
\journal{Elsevier}

\begin{document}

\begin{frontmatter}


\title{Implementation of Survivor Detection Strategies Using Drones}


\author[label1]{Sarthak J. Shetty\corref{cor1}}
\author[label2]{Rahul Ravichandran}
\author[label1]{Lima Agnel Tony}
\author[label3]{N. Sai Abhinay}
\author[label3]{Kaushik Das}
\author[label1]{Debasish Ghose\corref{cor1}}
\cortext[cor1]{E-mail: (sarthakjs + dghose)@iisc.ac.in}
\address[label1]{Department of Aerospace Engineering, Indian Institute of Science, Bengaluru, India}
\address[label2]{Aalborg University, Aalborg, Denmark}
\address[label3]{Tata Consultancy Services - Innovation Lab, Bangalore, India}

\begin{abstract}
\noindent Survivors stranded during floods tend to seek refuge on dry land. It is important to search for these survivors and help them reach safety as quickly as possible. The terrain in such situations however, is heavily damaged and restricts the movement of emergency personnel towards these survivors. Therefore, it is advantageous to utilize Unmanned Aerial Vehicles (UAVs) in cooperation with on-ground first-responders to aid search and rescue efforts. 

In this article we demonstrate an implementation and improvement of the “weight-based” path planning algorithm using an off-the-shelf UAV. The coordinates of the survivor and their heading is reported by an on-ground observer to the UAV to generate a weighted map of the surroundings for exploration. Each coordinate in the map is assigned a weight which dictates the priority of exploration. These waypoints are then sorted on the basis of their weights to arrive at an ordered list for exploration by the UAV.

We developed the model in MATLAB, followed by prototyping on Robot Operating System (ROS) using a 3DR Iris Quadcopter. We tested the model on an off-the-shelf UAV by utilizing the MAVROS and MAVLINK capabilities of ROS. During the implementation of the algorithm on the UAV, several additional factors such as unreliable GPS signals and limited field of view which could effect the performance of the model were in effect, despite which the algorithm performed fairly well.

We compared our model with conventional algorithms described in the literature, and showed that our implementation outperforms them.

\end{abstract}


\end{frontmatter}


\section{Introduction}\label{sec1}

\noindent Floods are one of the most devastating natural disasters, causing significant damage to property and human life. In order to speed up search and rescue efforts in the aftermath of such situations, a number of technological solutions have been proposed \cite{HumanBodyDetection}, \cite{DynamicRouting}. 

\medskip

\noindent Given their rapid commercialization, Unmanned Aerial Vehicles (UAVs) in recent years have become a viable option \cite{Avalanche}, \cite{CameraBased}, \cite{HelpFromSky} in the aftermath of disasters, serving as an aid in search and rescue operations.

\medskip

\noindent UAVs have been utilized in such scenarios given their ability to traverse large areas and carry out effective reconnaissance. UAVs can also be equipped with a variety of sensors to gather different modalities of data, ranging from temperature to visual information \cite{Survey}.

\medskip

\noindent These capabilities make the usage of UAVs particularly advantageous during, and in the aftermath of nuclear disasters \cite{Hawk}, hurricanes \cite{Hurricane} and forest fires \cite{Forest}.

\medskip

\noindent Path-planning algorithms, such as lawn-mower search, are employed to look for survivors in such scenarios. However, conventional algorithms make no use of prior information, gathered by on-site emergency personnel, to guide their decision making processes while searching for survivors. 

\medskip

\noindent In realistic scenarios, this information is of vital importance, especially while predicting a survivor's possible location in fast moving flood waters.  Therefore, there is a need for an exploration algorithm that utilizes the survivor's coordinates and direction.

\medskip

\noindent Hence, we propose a novel search algorithm that employs the survivor's information, as gathered by an on-ground observer, to prioritize waypoints for exploration by the UAV.

\medskip

\noindent In this chapter, we demonstrate a modification and implementation of the proposed "weight-based" model \cite{Weight}, where a survivor's location and heading, as observed by an on-ground emergency personnel, is utilized to generate a probabilistic map of the survivor's location.

\medskip

\noindent This probabilistic map is used to guide the UAV towards probable survivor locations by deferentially prioritizing waypoints, conditional to their relative heading to the survivor's direction.

\medskip

\noindent To present the objectives, methodology and results, this chapter is structured with the following sub-sections: 1.1 Introduction, 1.2 Modified Weight-Based Exploration, 1.3 Simulations, 1.4 Implementation, 1.5 Results, 1.6 Conclusion, 1.7 Future Work and Acknowledgements.

\subsection{Related Work}

\noindent Conventional planning algorithms for search and rescue, such as lawn-mower exploration, do not make use of prior information as reported by on-ground observers. The Weight-Based algorithm \cite{Weight} however, utilizes the survivor's coordinates and heading to generate a prioritized list of coordinates for exploration.

\medskip

\noindent Referencing Figure 1.3, the direction of the survivor's heading receives the highest priority of exploration, the quadrants on either side receiving the next priority and the direction opposite to the heading receives the least priority.

\begin{table}[h!]
\caption{Parameters of Monte-Carlo Simulations \cite{Weight}}
\centering
\begin{tabular}{|l|l|}
\hline
\textbf{Parameter}     & \textbf{Value}    \\ \hline
Environment Size       & 600 m x 600 m     \\ \hline
Survivor Speed         & 0.6 m/s and 2 m/s \\ \hline
UAV Max Speed          & 12 m/s            \\ \hline
UAV Search Radius      & 18 m x 18 m       \\ \hline
UAV Camera FOV         & \ang{45}               \\ \hline
UAV Flight Time        & 30 minutes        \\ \hline
Height of UAV          & 9 m               \\ \hline
Observer Search Radius & 30 m              \\ \hline
Number of observers    & 30                \\ \hline
Observer Positions     & Randomized        \\ \hline
\end{tabular}
\end{table}

\noindent We deployed the algorithm described, initially on MATLAB to test the veracity of the model \cite{Weight}. In order to test the model's capabilities against a standard lawn-mower search pattern, we carried out Monte-Carlo testing \cite{Weight} using a set of 10 parameters, from Table 1.1. 

\medskip

\noindent From the Monte-Carlo simulations performed in \cite{Weight}, we showed that in a simulation only environment, the Weight-Based method outperformed the standard lawn-mower and a weighted lawn-mower search methods. The results from the simulations in \cite{Weight} are shown Figure 1.1 and Figure 1.2.

\setlength{\fboxsep}{1.3pt}%
\setlength{\fboxrule}{1pt}%
\begin{figure}[h!]
    \centering
    \fbox{\includegraphics[scale=0.35]{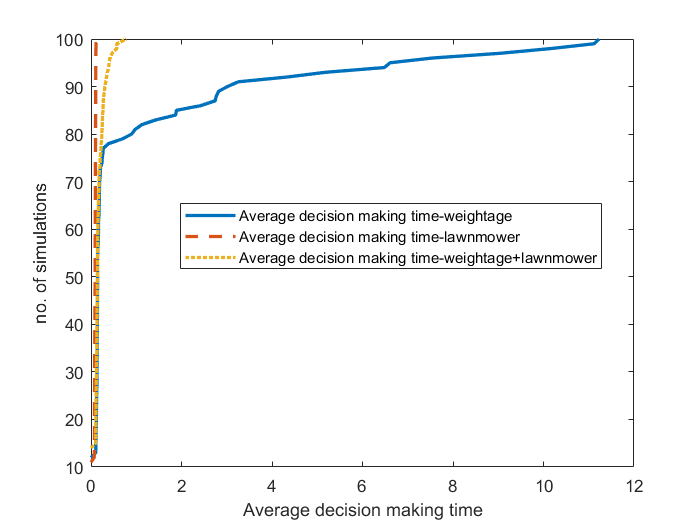}}
    \caption{Number of Simulations vs Average Decision Making Time for Search Algorithms \cite{Weight}}
    \label{fig:Monte1}
\end{figure}

\setlength{\fboxsep}{1.3pt}%
\setlength{\fboxrule}{1pt}%
\begin{figure}[h!]
    \centering
    \fbox{\includegraphics[scale=0.46]{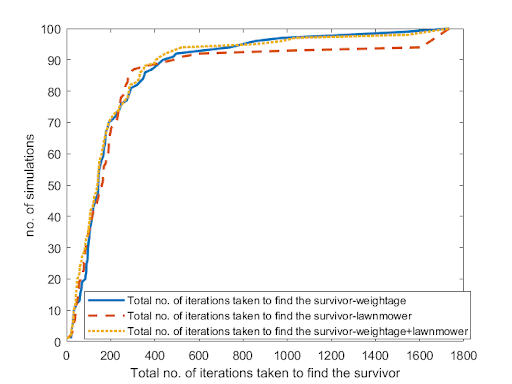}}
    \caption{Number of Simulations vs Number of Iterations Required to Find the Survivor \cite{Weight}}
    \label{fig:Monte2}
\end{figure}

\noindent In this chapter, we build upon this previous work \cite{Weight} and show that our model outperforms the conventional lawn-mower search both in ROS simulations and in real-life testing using an off-the-shelf UAV.

\medskip

\noindent The MATLAB environment was also used to perform the Monte-Carlo simulations presented under Section 1.1.1, to compare the decision making time and the time to reach the survivor against two existing search algorithms, the lawn-mower method and the probability based search.

\medskip

\noindent We then tested the model on Robot Operating System (ROS) in order to test the model's viability and robustness. After extensive testing in simulations we tested the model on an off-the-shelf UAV using ROS' compatibility with the Pixhawk flight controller hardware.

\medskip

\noindent We utilized the MAVROS capability of MAVLINK to communicate with the Pixhawk on-board the UAV. The physical set-up to carry out testing has been discussed in detail under Section 1.3.

\section{Modified Weight-Based Exploration}\label{subsec1}

The "weight-based" algorithm \cite{Weight} is a novel path-planning algorithm that generates a prioritized map of the survivor's possible location. This probabilistic model generates a prioritized list of waypoints by differentially assigning weights to each waypoint, depending on their location relative to the survivor's heading.

\noindent In Figure 1.1:

\begin{enumerate}
    \item The \textbf{green} path represents the lawn-mower trajectory of the UAV before being called by the observer to a survivor's last known coordinates.
    \item The \textbf{red} dot and accompanying blue dotted line represents the survivor's last known coordinates and direction as relayed to the UAV by the survivor.
    \item The \textbf{blue} path represents the trajectory of the UAV after the weight-based exploration has been triggered.
    \item The \textbf{numbers (1, 2, 3, 4)} at the corner of each quadrant represents their exploration priority during the weighted exploration.
\end{enumerate}

\begin{figure}[h]
    \centering   \includegraphics[scale=0.31]{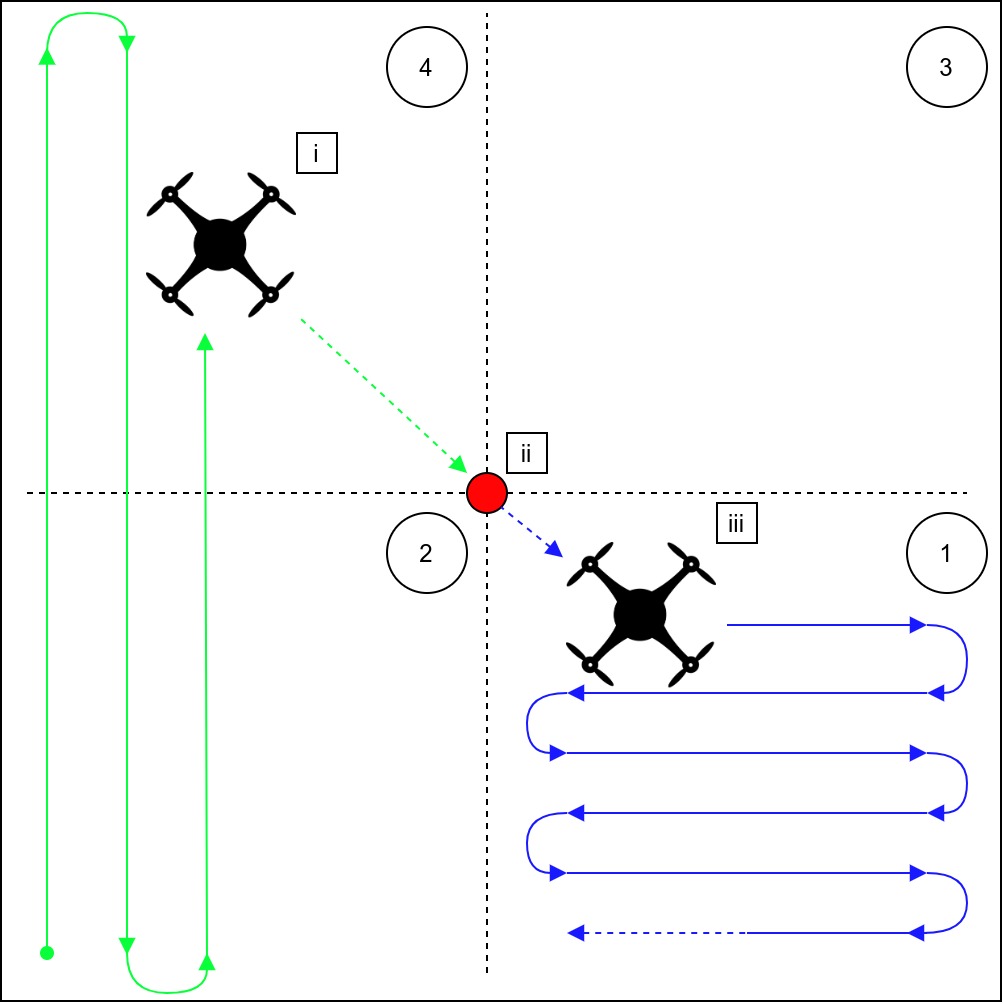}
    \caption{Representation of Weight-Based Exploration}
    \label{fig:Figure1}
\end{figure}

\subsection{Model Description}

\noindent In the model, we assume that the observer has a fixed radius of surveillance. If a survivor enters this surveillance region, the observer reports the survivor's information to the UAV. This information reported by the observer contains the survivor's coordinates and their heading.

\medskip

\noindent Once the UAV receives this information, it breaks away from the lawn-mower pattern of search and heads over to the coordinates of the survivor as reported by the observer.

\medskip

\noindent The UAV utilizes the survivor's information to generate the aforementioned weighted map of the surroundings. This approach is probabilistic in nature given that coordinates are assigned weights relative to their bearing from the survivor's reported direction.

\medskip

\noindent The weighted exploration utilizes the observer's information to create a prioritized list of waypoints, based on the following order:

\begin{enumerate}
    \item The survivor's last known coordinates receive the highest weight, ensuring that the UAV explores it first, before moving on to the rest of the coordinates.
    \item Coordinates that lie along the direction of the survivor's heading receive the highest weights.
    \item The directions on either side of the survivors heading have equal priority of exploration. If the heading is equally aligned to adjacent quadrants, we assume a higher priority to the left quadrant.
    \item Coordinates lying in direction opposite to the survivor's heading receive the lowest weights.
\end{enumerate}

\medskip

\noindent This iterative differential assignment of weights to the coordinates based on their proximity to the survivor's location and heading is utilized to rank them and arrive at an ordered list of waypoints, where the waypoints with the highest weights are explored first by the UAV, as they are the most probable locations where the survivor might be present.

\medskip

\noindent This ordered list is conveyed to the UAV for sequential exploration of the region. We utilize the MAVROS capability of ROS to achieve a sequential delivery of ordered waypoints to the UAV.

\subsubsection{\textbf{Weight Calculation}}

\noindent In \cite{Weight} the weights assigned to the quadrants were calculated heuristically. In large environments, a condition may arise when equal weights may be assigned to coordinates in two different quadrants. Such a condition, as observed in simulations as well, causes erratic movement of the UAV while moving from one waypoint to another.

\medskip

\noindent To prevent such a condition from arising, we devise a set of equations which take into consideration the size of the environment and the survivor's position to generate a set of weights which will be used during iterative assignment to the coordinates.

\medskip

\noindent Let the weights assigned to the four quadrants be denoted as \emph{W\textsubscript{1}}, \emph{W\textsubscript{2}}, \emph{W\textsubscript{3}}, \emph{W\textsubscript{4}} and \emph{W\textsubscript{5}} be the weight assigned to the survivor's last reported coordinate.

\medskip

\noindent In accordance to priority of quadrants described in the Model Description in Section 1.1.3:

\begin{enumerate}
    \item Let \emph{W\textsubscript{1}} be the weight assigned to the quadrant along the survivor's direction of heading. 
    \item Let \emph{W\textsubscript{2}} and \emph{W\textsubscript{3}} be the weights assigned to quadrants on left and right of the survivor's heading respectively. 
    \item Let \emph{W\textsubscript{4}} be the weight assigned to the direction opposite to the survivor's heading.
    \item Let \emph{W\textsubscript{5}} be the weight assigned to the survivor's last known coordinates.
\end{enumerate}

\noindent Given these conditions, assuming the heading to be equal alignment of the heading to either adjacent quadrants, the order of weights is as follows:

\begin{equation} \label{eq:1}
\centerline{\emph{W1} $>$ \emph{W2} $>$ \emph{W3} $>$ \emph{W4}}
\end{equation}

\noindent We define a set for each quadrant, spanning the least possible weight (\emph{W}\textsubscript{i}) to the maximum possible weight attained by a coordinate lying in that coordinate (\emph{W}\textsubscript{i max}).

\noindent Such a set can be denoted as:

\centerline{[\emph{W\textsubscript{i}}, \emph{W\textsubscript{n max}}]}

\noindent \emph{W\textsubscript{i max}} can be defined as:

\centerline{\emph{W\textsubscript{i max}} = \emph{N} * \emph{W\textsubscript{i}}}

\medskip

\noindent Here, \emph{N} is the maximum number of iterations required to move from the survivor's coordinates to the boundary of the environment (X\textsubscript{e}, Y\textsubscript{e}).
\noindent It can be calculated as:

\centerline{\emph{N} = X\textsubscript{e} - X\textsubscript{s}}
\centerline{or} 
\centerline{\emph{N} = Y\textsubscript{e} - Y\textsubscript{s}}

\noindent whichever is greater.

\medskip

\noindent From the inequality (\ref{eq:1}), the necessary condition to prevent common weights among quadrants is:

\begin{equation} \label{eq:2}
\centerline{\emph{W\textsubscript{i}} $>$ \emph{W\textsubscript{(i-1) max}}}
\end{equation}

\noindent The minimum condition for (\ref{eq:2}) to be satisfied, assuming \emph{W\textsubscript{i}} to be integers, is:

\begin{equation}\label{eq:3}
\centerline{\emph{W\textsubscript{i}} = \emph{W\textsubscript{(i-1) max}} + 1}
\end{equation}

\noindent From (\ref{eq:1}) and (\ref{eq:2}) we arrive at the following set of equations for the corresponding weights:

\begin{equation} \label{eq:4}
\emph{W\textsubscript{1}} = ((\emph{W\textsubscript{4}}  N^3) + N + N^2 + N^3)
\end{equation}
\begin{equation} \label{eq:5}
\emph{W\textsubscript{2}} = (\emph{W\textsubscript{1}} - N) / N
\end{equation}
\begin{equation} \label{eq:6}
\emph{W\textsubscript{3}} = ((\emph{W\textsubscript{1}} - N) - N^2)) / N^2
\end{equation}
\begin{equation} \label{eq:7}
\emph{W\textsubscript{5}} = ((\emph{W\textsubscript{1}}N) + 1)
\end{equation}

\noindent Equation (\ref{eq:3}) assigns a weight to the survivor's coordinate that is greater than the highest possible weight assigned along the survivor's heading. 

\setlength{\fboxsep}{1.3pt}%
\setlength{\fboxrule}{1pt}%
\begin{figure}[h!]
    \centering
    \fbox{\includegraphics[scale=0.80]{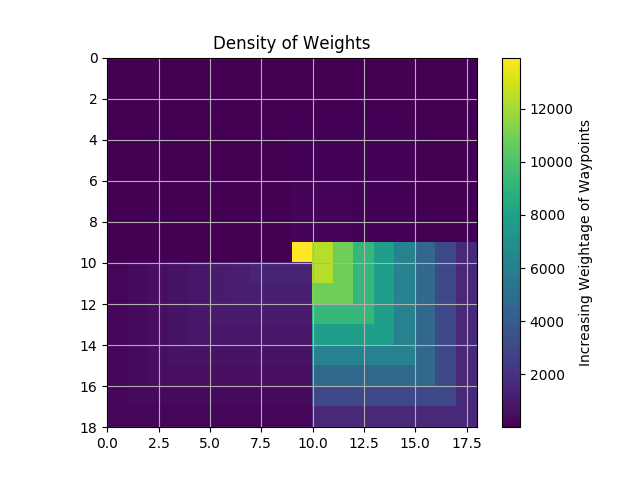}}
    \caption{Weight Density Across the Simulation Environment}%
\end{figure}

\noindent Here, \emph{W\textsubscript{4}}, the weight in the direction opposite to the survivor's heading, is assumed to be 1, the lowest non-negative integer value that can be assigned as a weight.

\setlength{\fboxsep}{1.3pt}%
\setlength{\fboxrule}{1pt}%
\begin{figure}[h!]
    \centering
    \fbox{\includegraphics[scale=0.80]{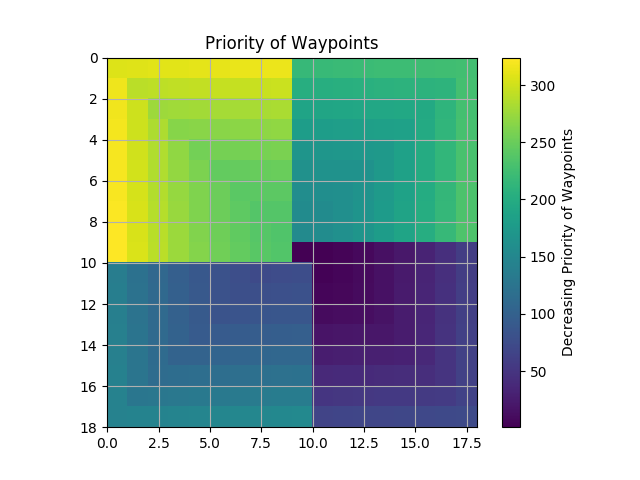}}
    \caption{Priority of Exploration of Waypoints}%
\end{figure}

\noindent In accordance with equation (\ref{eq:4}), (\ref{eq:5}),  (\ref{eq:6}), and (\ref{eq:7}) a color gradient can be observed across quadrants due to differential assignment of the weights.

\medskip

\noindent This map, of differentially weighted coordinates, is used to prioritize and generate a list of waypoints used by the UAV for path-planning in the environment. Figure 1.5 represents the exploration map after prioritization.

\section{Simulations}

\noindent Before the implementation of the algorithm on the UAV, we tested the model on MATLAB in \cite{Weight}, followed by simulations on Robot Operating System (ROS), using a standard 3DR Iris Quadcopter. Both of these simulations environments are discussed in the sections below.

\subsection{MATLAB Simulations}\label{subsec2}

\noindent The initial simulations to test the veracity of the algorithm were performed on MATLAB \cite{Weight}. The environment on MATLAB is shown in a series of sub-figures in Figure 1.6. 

\medskip

\noindent The environment shown in Figure 1.4 was also used to arrive at the Monte-Carlo simulation results as presented in Section 1.1.1.

\bigbreak

\noindent In the 600 m x 600 m MATLAB environment shown in Figure 1.6:
\begin{enumerate}
    \item Multiple observers are assumed, represented by \textbf{blue} circles. The observers are assumed to have a fixed radius of observation.
    \item The UAV is described as a \textbf{pink} box with a trailing pink dotted trajectory.
    \item The survivor is represented by a \textbf{red} asterisk, with a red trajectory.
\end{enumerate}

\medskip

\setlength{\fboxsep}{1.3pt}%
\setlength{\fboxrule}{1pt}%
\begin{figure}[h]
    \centering
    \subfloat{\fbox{{\includegraphics[scale=0.15]{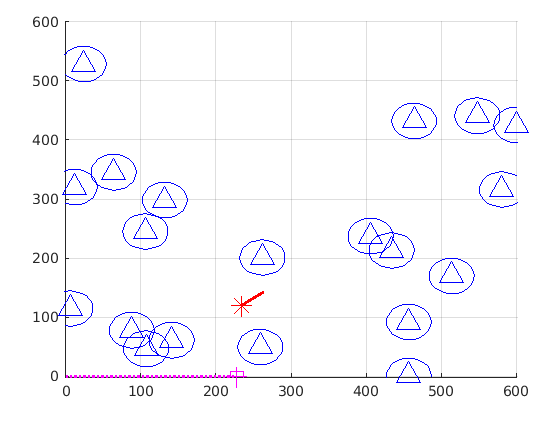} }}}%
    \qquad
    \subfloat{\fbox{{\includegraphics[scale=0.15]{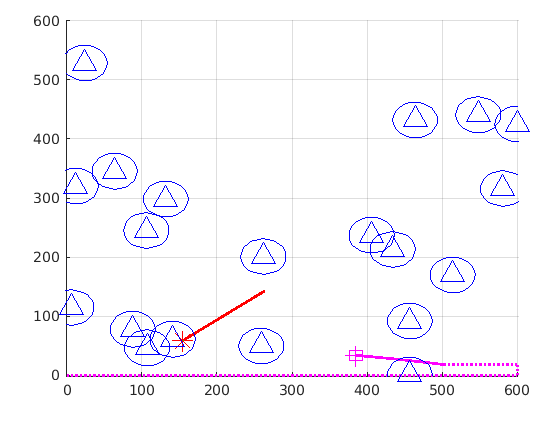} }}}%
    \qquad
    \subfloat{\fbox{{\includegraphics[scale=0.15]{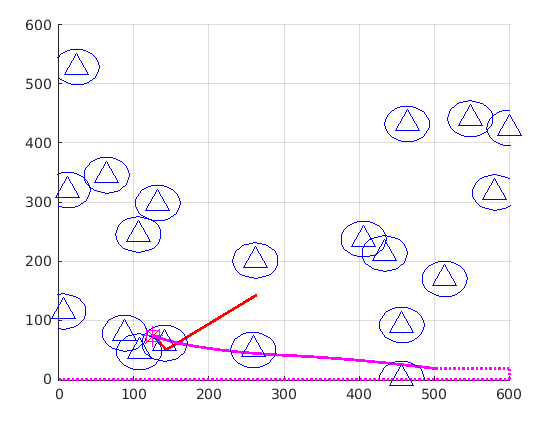} }}}%
    \caption{MATLAB Simulations of Model \cite{Weight}}%
    \label{fig:MATLAB_SIMULATION}%
\end{figure}

\noindent In Figure 1.4:
\begin{enumerate}
    \item In the \textbf{first} sub-figure, the UAV is initially executing the lawn-mower pattern of search and the survivor is set to a random trajectory.
    \item In the \textbf{second} sub-figure, the survivor breaches the radius of one of the observers. The UAV breaks away from lawn-mower search and begins executing the weight-based search, utilizing the heading and coordinates of the survivor.
    \item Through the weight-based search, the UAV eventually catches up to the survivor in the \textbf{third} sub-figure.
\end{enumerate}

\medskip

\noindent For the Monte-Carlo simulations on MATLAB, the variables in Table 1.1 were considered, including varying number of observers with randomized positions, and variable survivor trajectories as well.

\subsection{Robot Operating System Simulations}\label{subsec3}

The model was ported from MATLAB to ROS prior to testing on hardware, given the compatibility of the hardware systems with the ROS framework.  The code was written primarily in C++. As mentioned, in order to communicate with the Pixhawk 1 autopilot on-board the simulated Iris Drone and the physical drone, the MAVLINK and MAVROS capabilities of ROS were used.

\setlength{\fboxrule}{1pt}%
\begin{figure}[h!]
    \centering
    \subfloat{\fbox{{\includegraphics[scale=0.42]{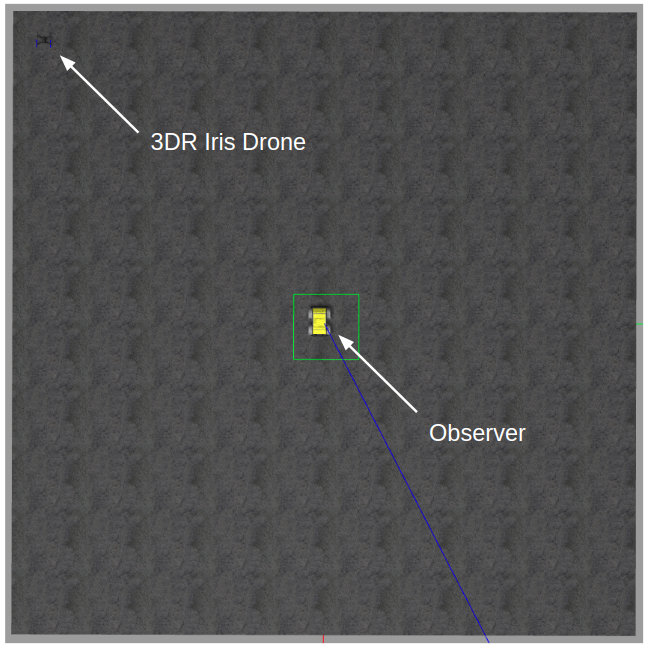} }}}%
    \caption{ROS Environment}
    \label{fig:ROSEnvironment}%
\end{figure}

\noindent A 3DR Iris Quadcopter was used to prototype the model on ROS, with an on-board Pixhawk autopilot board. This simulation configuration was selected because of the semblance to our real drone, ensuring compatibility of the ROS and C++ modules written for the two systems.

\medskip

\noindent Figure 1.5 represents the prototyping ROS environment. The environment is assumed to be 20 m x 20 m. The Iris Quadcopter is located at the top left. A single observer is assumed, located at the (10, 10) coordinate of the environment, represented by a ClearPath Husky.

\medskip

\noindent During initialization of the ROS environment the controller generates a local coordinate system with the UAV localized at the origin. We utilize this local coordinate system to generate the waypoints for navigation and for the UAV's path planning.

\medskip
\noindent ROS also provides an interface to debug the model prior to hardware testing, through tools such as RViz, rqt\_graph and ROS Bags. In order to log data such as the positions of the UAV and the survivor for visualizations, we run nodes to subscribe to these parameters.

\section{Implementation}

\noindent For the implementation of the model on the UAV, as mentioned, we use the MAVLINK and MAVROS capabilities of ROS to communicate with the on-board autopilot. In our case the on-board autopilot is a Pixhawk 1. The physical setup is as shown in Figure 1.8.

\setlength{\fboxsep}{1.3pt}%
\setlength{\fboxrule}{1pt}%
\begin{figure}[h]
    \centering
    \fbox{\includegraphics[scale=0.16]{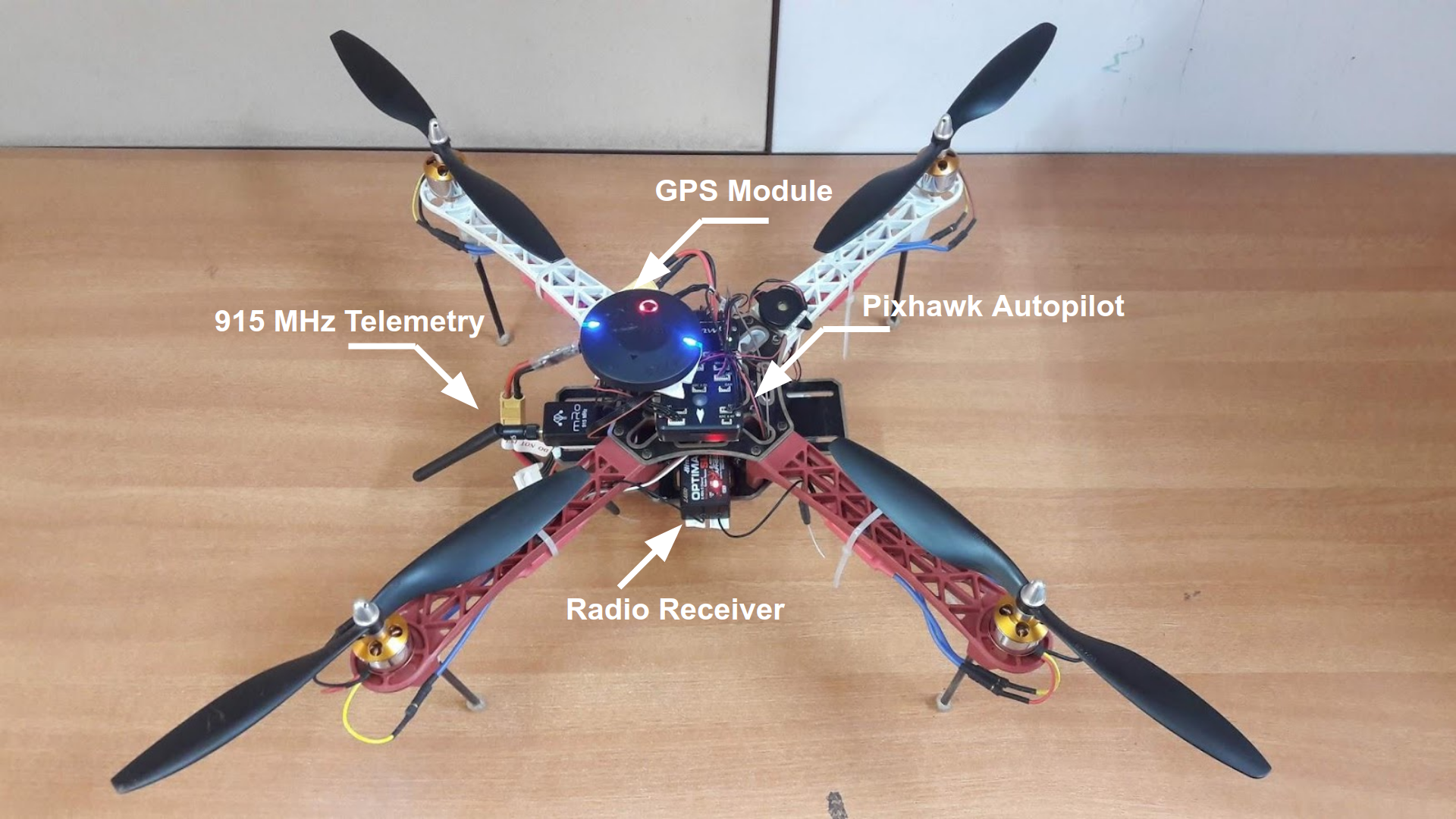}}
    \label{fig:Hardware}
    \caption{Hardware setup for physical testing}
\end{figure}

\medskip

\noindent The Pixhawk enables interfacing with the ROS code that runs the UAV during simulations, therefore ensuring similar performance and compatibility with the C++ code.

\setlength{\fboxsep}{1.3pt}%
\setlength{\fboxrule}{1pt}%
\begin{figure}[h!]
    \centering
    \fbox{\includegraphics[scale=0.18]{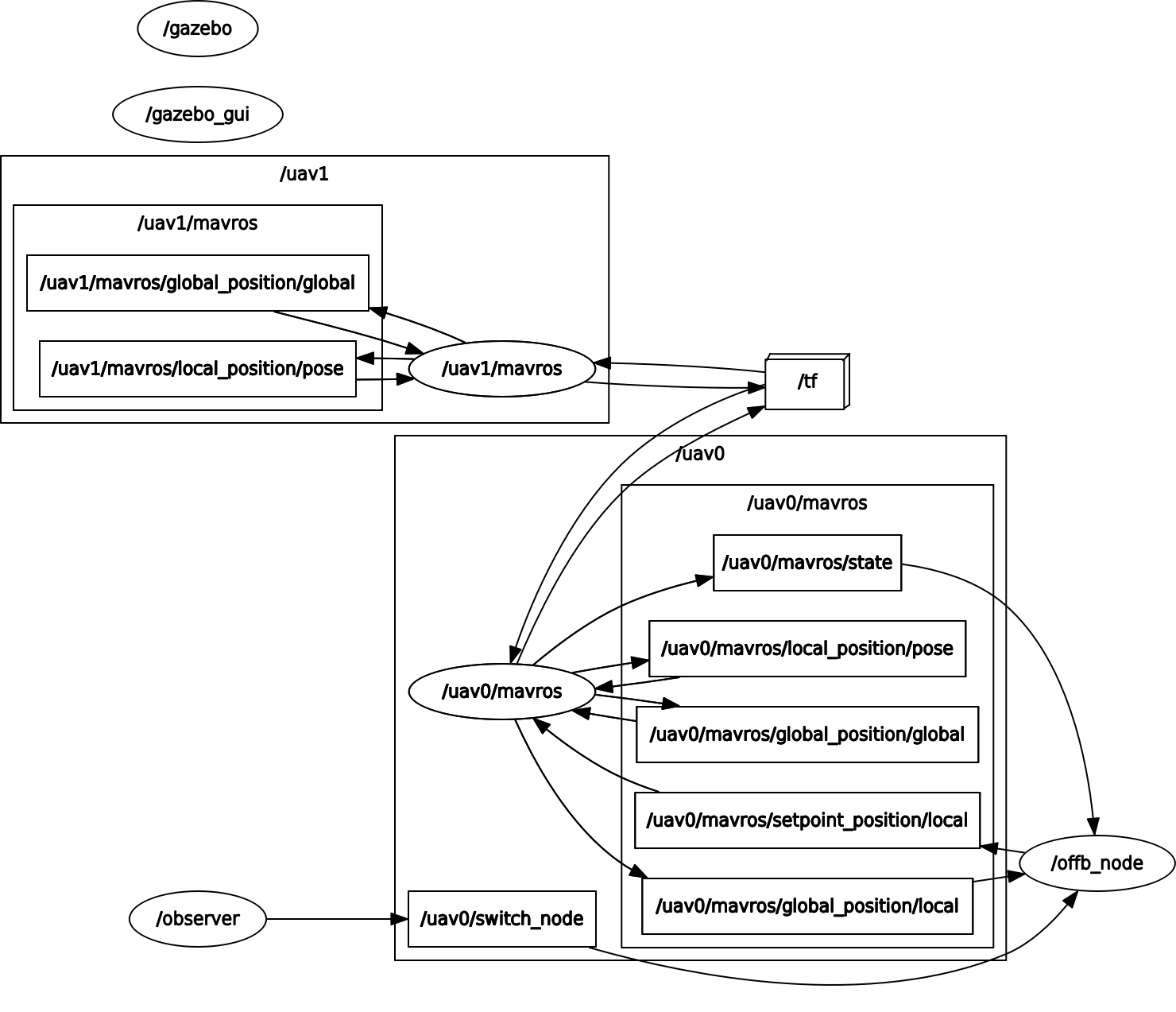}}
    \label{fig:ROSGraph}
    \caption{ROS Graph of interacting nodes during simulation}
\end{figure}

\medskip

\noindent Figure 1.9 represents the ROS Graph, which can be used to visualize the various nodes transacting topics amongst each other during the ROS simulation.

\section{Results}

\noindent We use the time taken by the UAV to find the survivor as a metric of each model's performance. We assume an environment of 20 m x 20 m for the ROS simulations and a testing environment of 10 m x 10 m for physical testing. The variation in the UAV's X, Y and Z coordinates are presented as well. The survivor velocity \emph{V\textsubscript{s}} is set at 0.6 m/s for simulation and 0.3 m/s for physical testing.

\medskip

\noindent To plot the figures shown in each of the following sections, we have created Python scripts subscribing to the MAVROS \texttt{/mavros/global\_position} topic, as visualized in Figure 1.9. The data from this topic is cleaned and plotted using the \texttt{matplotlib} \cite{matplotlib} and \texttt{numpy} \cite{numpy} libraries for Python.

\subsection{ROS Results}

\subsubsection{\textbf{Weight-Based Exploration}}

\noindent The survivor is assumed to move linearly with a velocity \emph{V\textsubscript{s}} = 0.6 m/s. The total time taken for the weight-based survivor search to conclude is 213 seconds.

\medskip

\noindent As evident from figures 1.10, 1.11 and 1.12, the UAV initially investigating the environment using the lawn-mower search pattern breaks away to the last known location of the survivor at the (10, 10) coordinate.

\setlength{\fboxsep}{1.3pt}%
\setlength{\fboxrule}{1pt}%
\begin{figure}[h]
    \centering
    \fbox{\includegraphics[scale=0.37]{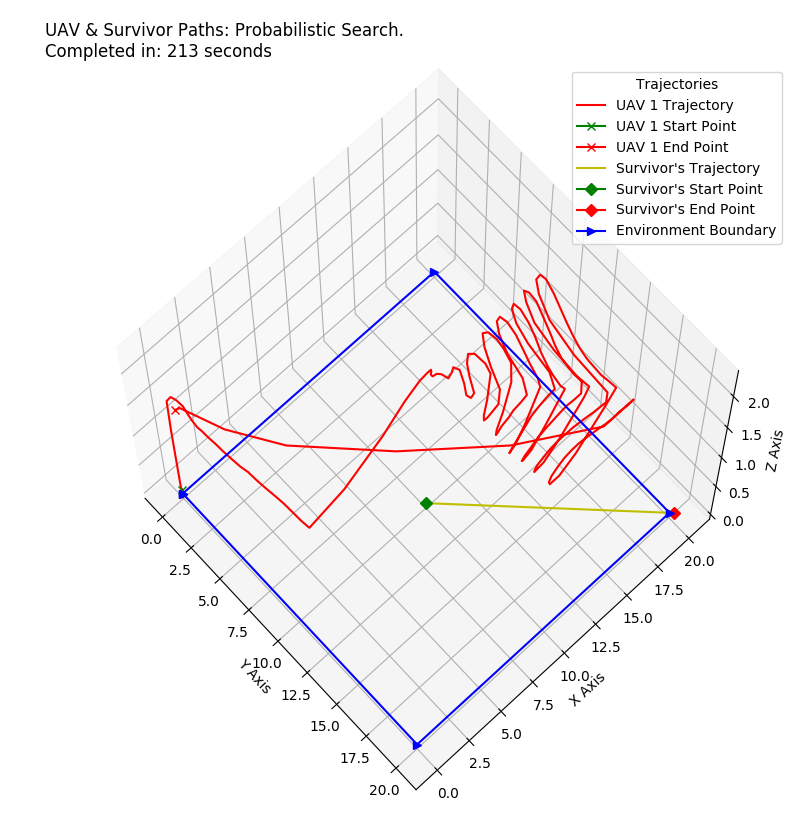}}
    \label{fig:06WeightTrajectory}
    \caption{Weight-Based Trajectory of UAV and Survivor \emph{V\textsubscript{s}} = 0.6 m/s}
\end{figure}

\setlength{\fboxsep}{1.3pt}%
\setlength{\fboxrule}{1pt}%
\begin{figure}[h]
    \centering 
    \subfloat{\fbox{{\includegraphics[scale=0.11]{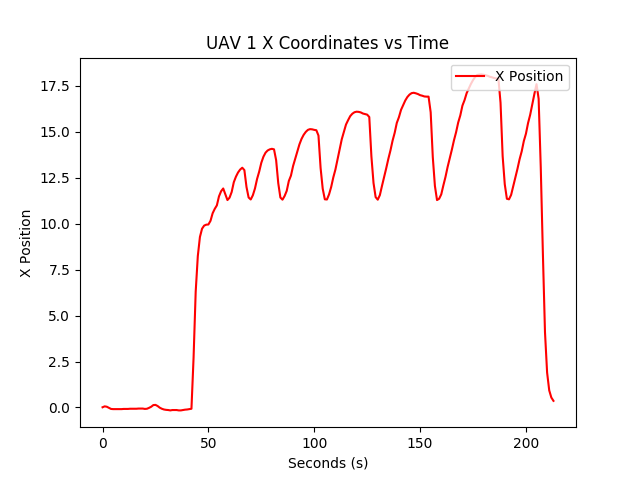} }}}%
    \qquad
    \subfloat{\fbox{{\includegraphics[scale=0.11]{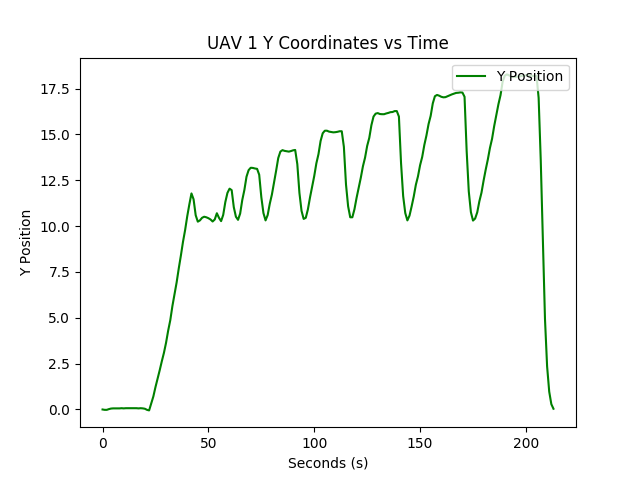} }}}%
    \qquad
    \subfloat{\fbox{{\includegraphics[scale=0.11]{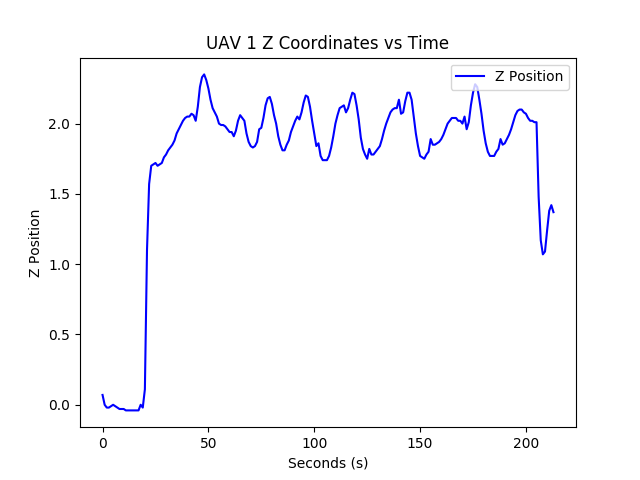} }}}%
    \caption{Variation in X, Y, Z Coordinates of UAV}%
    \label{fig:XYZUAV}%
\end{figure}
\setlength{\fboxsep}{1.3pt}%
\setlength{\fboxrule}{1pt}%
\begin{figure}[h!]
    \centering
    \subfloat{\fbox{{\includegraphics[scale=0.11]{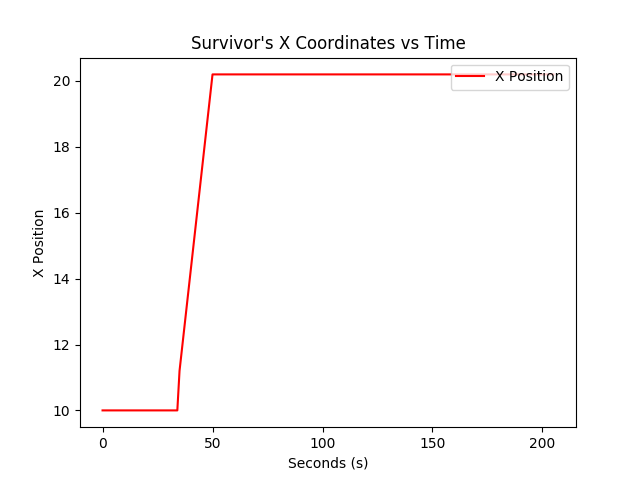} }}}%
    \qquad
    \subfloat{\fbox{{\includegraphics[scale=0.11]{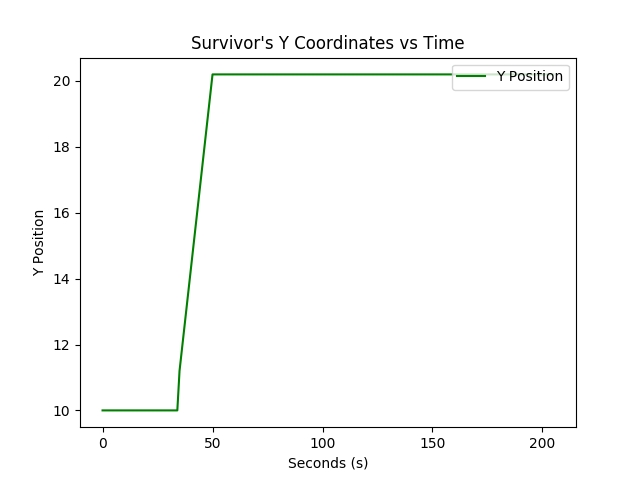} }}}%
    \qquad
    \subfloat{\fbox{{\includegraphics[scale=0.11]{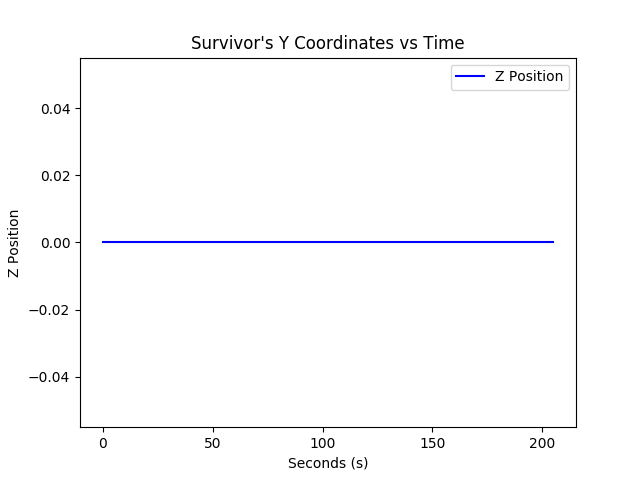} }}}%
    \caption{Variation in X, Y, Z Coordinates of Survivor \emph{V\textsubscript{s}} = 0.6 m/s}%
    \label{fig:XYZSurvivor}%
\end{figure}

\subsubsection{\textbf{Lawn-Mower Exploration}}

\noindent The survivor is assumed to move linearly with a velocity \emph{V\textsubscript{s}} = 0.6 m/s. The total time taken for the survivor search to conclude is 669 seconds. 

\medskip

\noindent From the simulations, it is apparent that the search time of the weight-based exploration outperforms the search time of the lawn-mower exploration by 215\%. 

\setlength{\fboxsep}{1.3pt}%
\setlength{\fboxrule}{1pt}%
\begin{figure}[h]
    \centering
    \fbox{\includegraphics[scale=0.36]{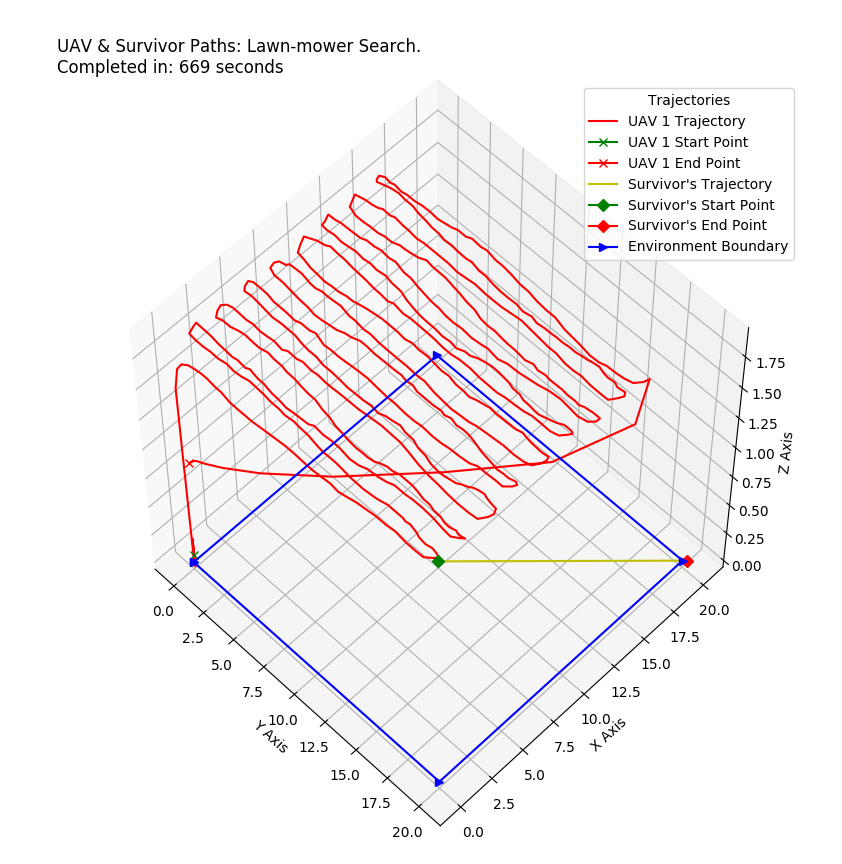}}
    \label{fig:06LMWeightTrajectory}
    \caption{Lawn-Mower Trajectory of UAV and Survivor \emph{V\textsubscript{s}} = 0.6 m/s}
\end{figure}
\setlength{\fboxsep}{1.3pt}%
\setlength{\fboxrule}{1pt}%
\begin{figure}[h!]
    \centering 
    \subfloat{\fbox{{\includegraphics[scale=0.11]{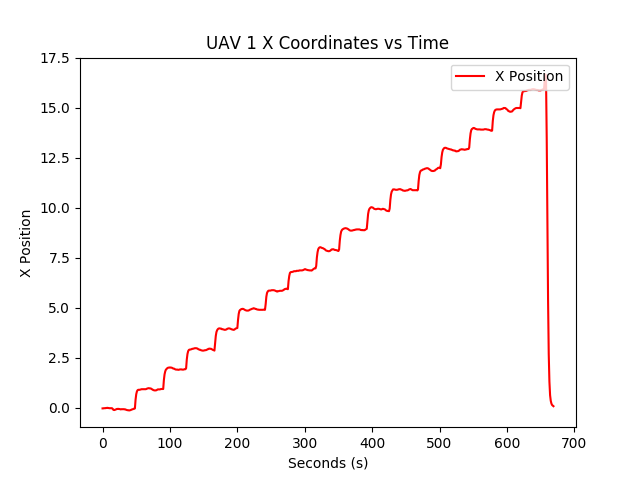} }}}%
    \qquad
    \subfloat{\fbox{{\includegraphics[scale=0.11]{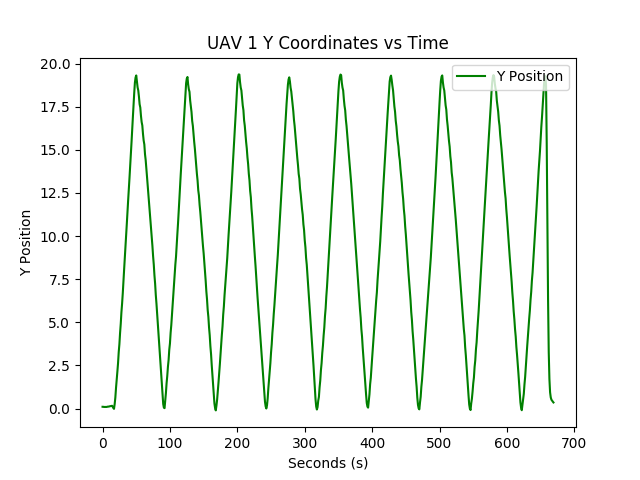} }}}%
    \qquad
    \subfloat{\fbox{{\includegraphics[scale=0.11]{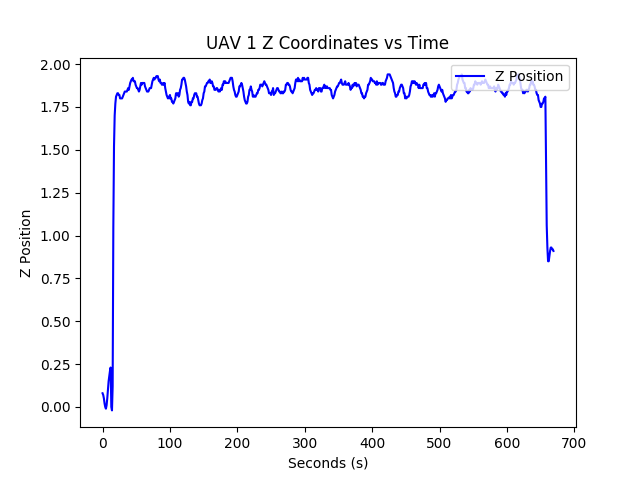} }}}%
    \caption{Variation in X, Y, Z Coordinates of UAV}%
    \label{fig:LMXYZUAV}%
\end{figure}
\setlength{\fboxsep}{1.3pt}%
\setlength{\fboxrule}{1pt}%
\begin{figure}[h!]
    \centering
    \subfloat{\fbox{{\includegraphics[scale=0.11]{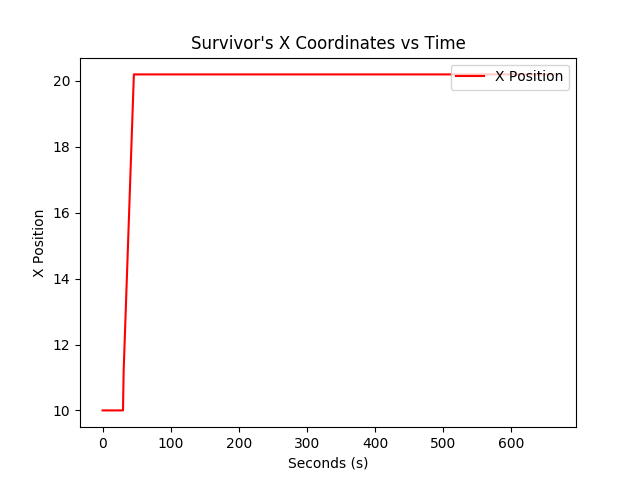} }}}%
    \qquad
    \subfloat{\fbox{{\includegraphics[scale=0.11]{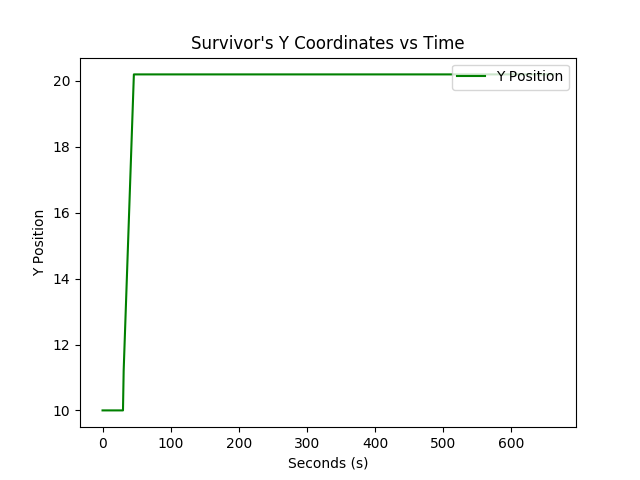} }}}%
    \qquad
    \subfloat{\fbox{{\includegraphics[scale=0.11]{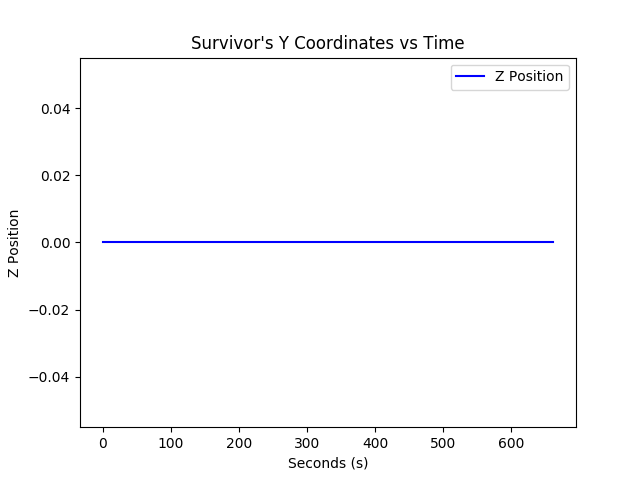} }}}%
    \caption{Variation in X, Y, Z Coordinates of Survivor \emph{V\textsubscript{s}} = 0.6 m/s}%
    \label{fig:LMXYZSurvivor}%
\end{figure}

\medskip

\noindent We have tabulated the search time for the two exploration strategies with varying survivor velocities (\emph{V\textsubscript{s}} = 0.6 m/s and \emph{V\textsubscript{s}} = 0.3 m/s) and two different environment sizes (20 m x 20 m and 18 m x 18 m). These results are presented in Table 1.2.

\begin{table}[h]
\caption{Search Times for Lawn-Mower and Weight-Based Exploration}
\centering
\begin{tabular}{|l|l|l|l|}
\hline
\textbf{Environment Size} ($m^2$)    & \textbf{\emph{V\textsubscript{s}}} (m/s)     & \textbf{\emph{T\textsubscript{L}}} (s) & \textbf{\emph{T\textsubscript{W}}} (s) \\ \hline
18  x 18          & 0.6  & 624    & 173 \\ \hline
20  x 20          & 0.6  & 669    & 213 \\ \hline
18  x 18          & 0.3  & 600    & 63  \\ \hline
20  x 20          & 0.3  & 663    & 66  \\ \hline
\end{tabular}
\end{table}

\noindent \emph{V\textsubscript{s}} - Survivor Velocity 
\\
\noindent \emph{T\textsubscript{L}} - Time taken to find survivor using lawn-mower search 
\\
\noindent \emph{T\textsubscript{W}} - Time taken to find survivor using weight-based search

\medskip

\noindent From Table 1.2 it is evident that our model comprising of weight-based exploration clearly outperforms, by nearly an order of magnitude, the standard lawn-mower search pattern, considering varied environment sizes and survivor velocities.

\medskip

\noindent As expected, the distance between the UAV and the observer at the time of invoking the \texttt{observer\_node} influences the time taken to reach the survivor. A larger search environment increases the time taken by the UAV to locate the survivor while following either the lawn-mower pattern or the weight-based algorithm, as evident from the difference between the search times for the 18 m x 18 m and 20 m x 20 m areas in Table 1.2.

\medskip 

\noindent Given similar environmental constraints, the search time for the weight-based method is significantly smaller than that for the lawn-mower pattern of search. Furthermore, for the weight-based method specifically, the search time is smaller for lower velocities as opposed to higher velocities. 

\medskip

\noindent This variation in the search time can be attributed to the fact that a higher velocity implies that the survivor moves away much more quickly from the last reported position, towards the boundaries of the observational environment.

\medskip

\noindent In the implementation sub-section, we show that the weight-based algorithm outperforms the lawn-mower method, despite unfavorable conditions arising due to strong winds and unreliable GPS signals.

\subsection{Implementation Results}
\noindent We carried out the physical tests in a 10 m x 10 m area. As described under Section 1.3, we use an off-the-shelf UAV with the components as described in Figure 1.8. The Pixhawk 1 autopilot on-board runs a MAVROS node that communicates with the on-ground system that runs the ROS model.

\medskip

\noindent The virtual survivor is a simple mathematical model which moves linearly in a prescribed direction with a velocity of 0.3 m/s. It is assumed to originate from the observer's position at (5, 5) in the 10 m x 10 m environment.

\medskip

\noindent Each sub-section below has the X-Y projection accompanying the 3D trajectory of the prototyping environment to provide a better perspective of the respective search patterns.

\subsubsection{\textbf{Weight-Based Exploration}}

\medskip

\setlength{\fboxsep}{1.3pt}%
\setlength{\fboxrule}{1pt}%
\begin{figure}[h!]
    \centering
    \subfloat{\fbox{{\includegraphics[scale=0.191]{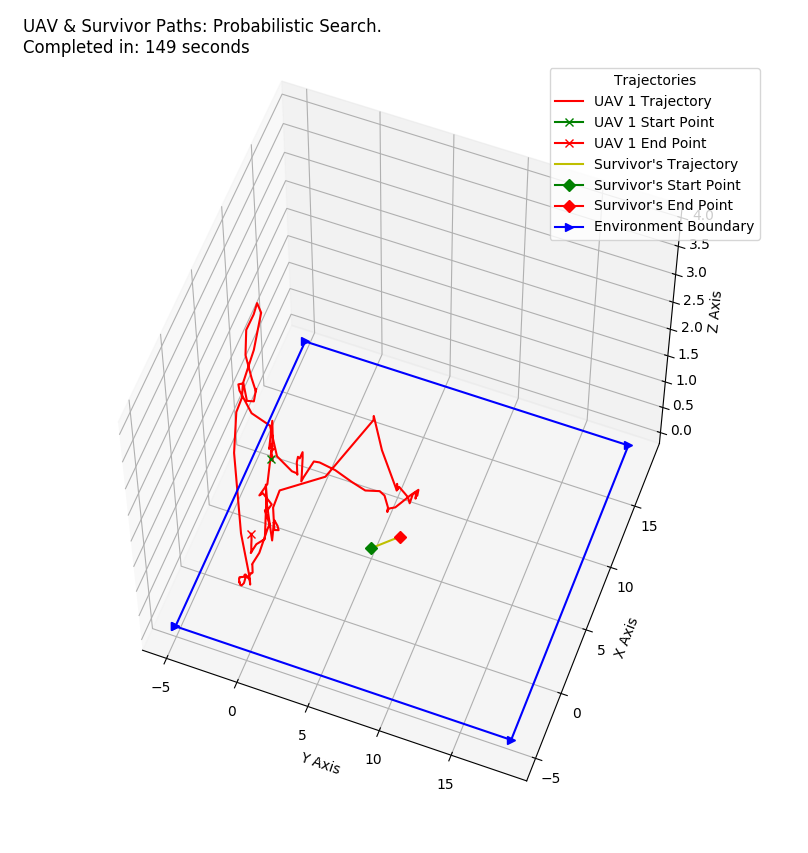}}}}%
    \qquad
    \subfloat{\fbox{{\includegraphics[scale=0.12]{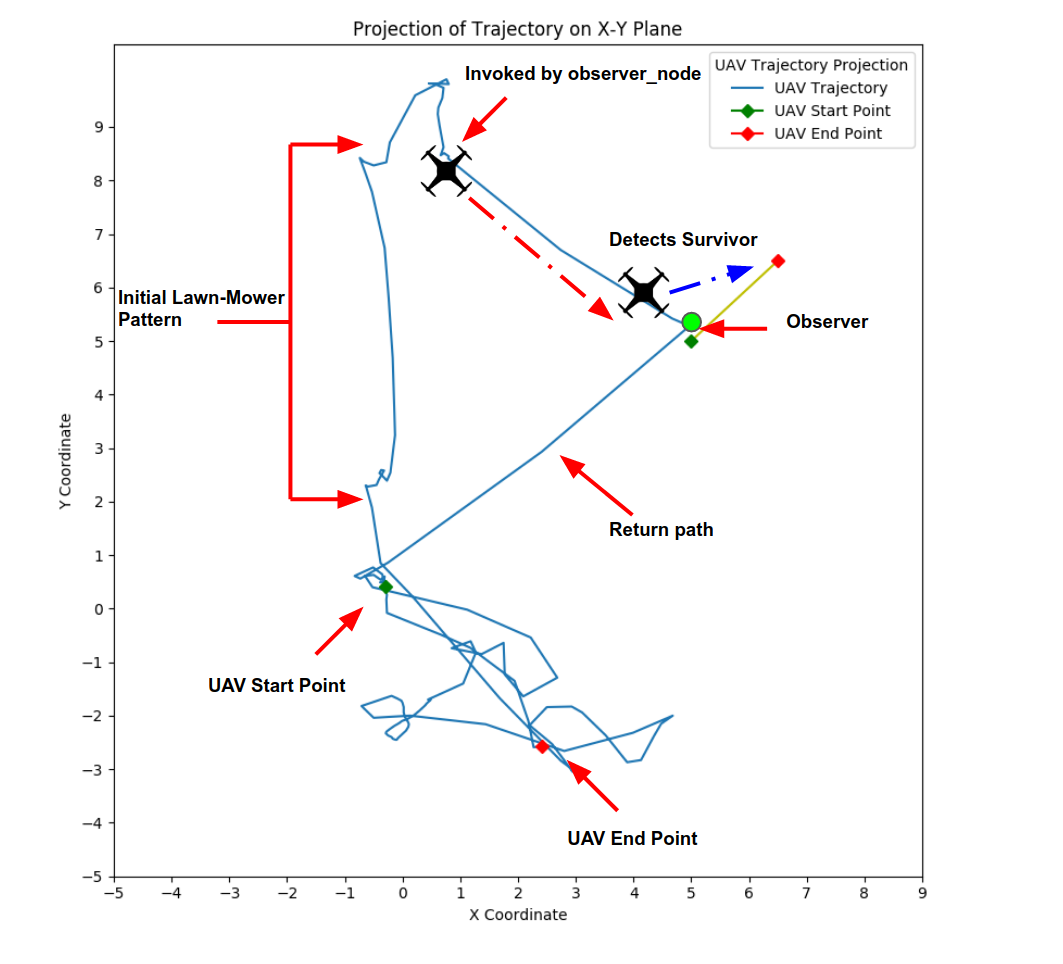}}}}%
    \caption{Trajectory and X-Y Projection of UAV and Survivor}%
    \label{fig:ProjectionPTLMXYZTrajectory}%
\end{figure}
\begin{figure}[h!]
    \centering
    \subfloat{\fbox{{\includegraphics[scale=0.15]{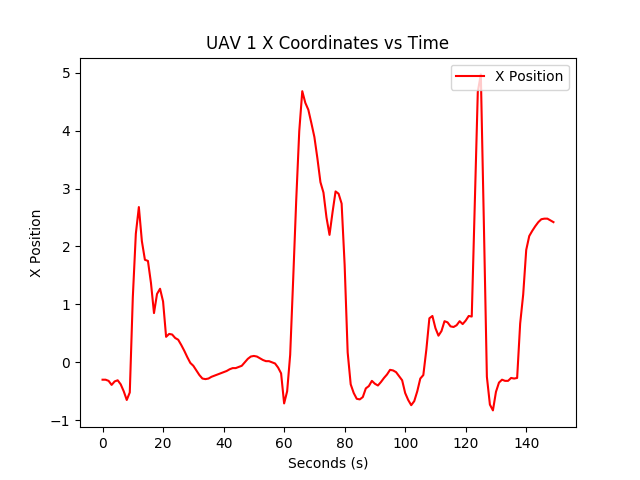} }}}%
    \qquad
    \subfloat{\fbox{{\includegraphics[scale=0.15]{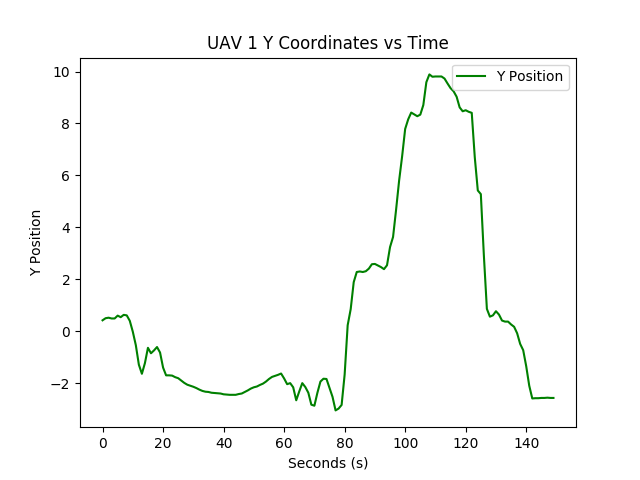} }}}%
    \qquad
    \subfloat{\fbox{{\includegraphics[scale=0.15]{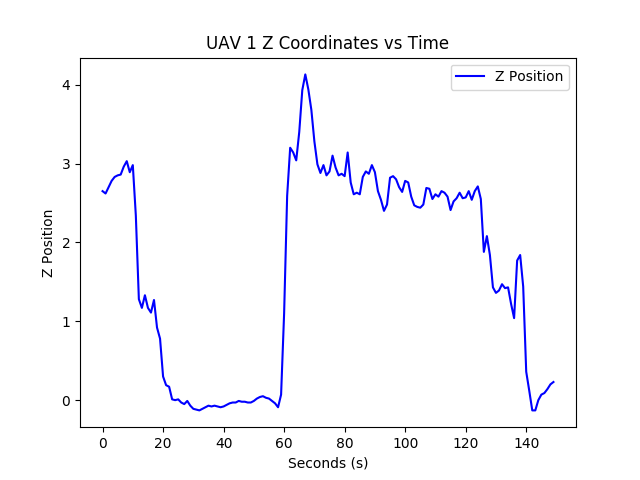} }}}%
    \caption{Variation in X, Y, Z Coordinates of Survivor \emph{V\textsubscript{s}} = 0.3 m/s}%
    \label{fig:PTLMXYZSurvivor}%
\end{figure}
\begin{figure}[h!]
    \centering
    \subfloat{\fbox{{\includegraphics[scale=0.15]{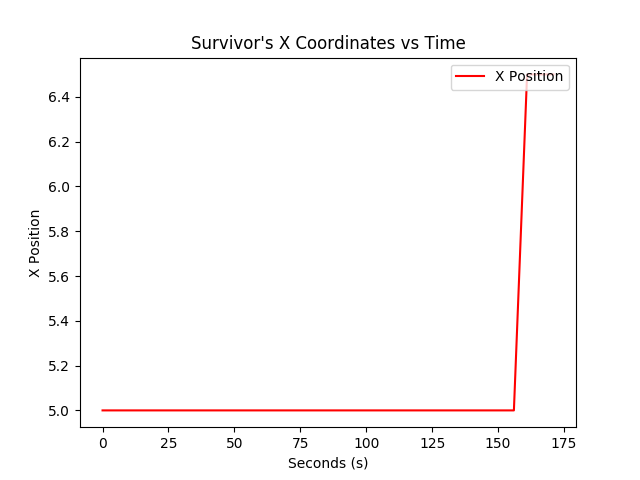} }}}%
    \qquad
    \subfloat{\fbox{{\includegraphics[scale=0.15]{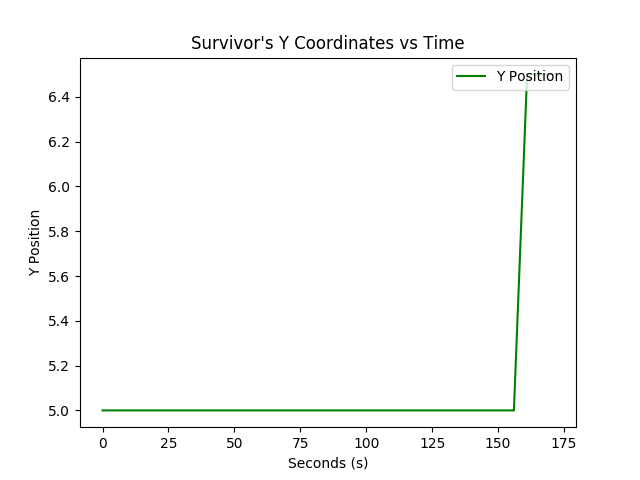} }}}%
    \qquad
    \subfloat{\fbox{{\includegraphics[scale=0.15]{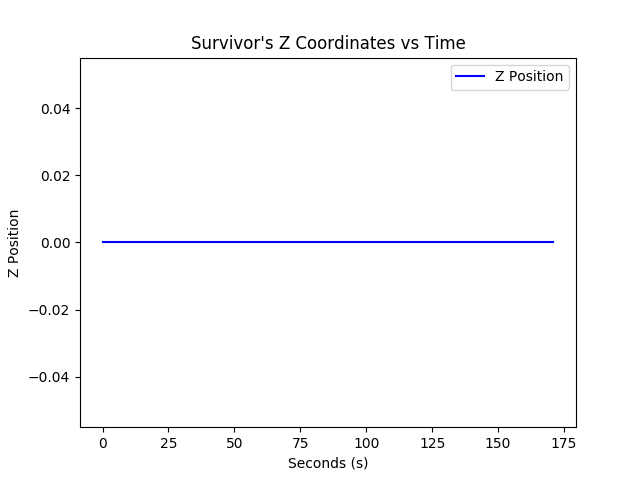} }}}%
    \caption{Variation in X, Y, Z Coordinates of UAV}%
    \label{fig:PTXYZUAV}%
\end{figure}

\noindent As observed from Figure 1.16, 1.17 and 1.18, the UAV successfully manages to execute the weight-based exploration and catches up to the survivor and returns back to the start coordinates with this information.

\medskip

\noindent The UAV takes 149 seconds to complete this operation. In the following section we compute the search time with the standard lawn-mower search pattern and compare the results of the two methods.

\subsubsection{\textbf{Lawn-Mower Exploration}}

\noindent For lawn-mower exploration as well, a 10 m x 10 m area is used for testing. The survivor is located at (5, 5) in the local coordinate system spawned by the UAV during initializing.

\setlength{\fboxsep}{1.3pt}%
\setlength{\fboxrule}{1pt}%
\begin{figure}[h!]
    \centering
    \subfloat{\fbox{{\includegraphics[scale=0.165]{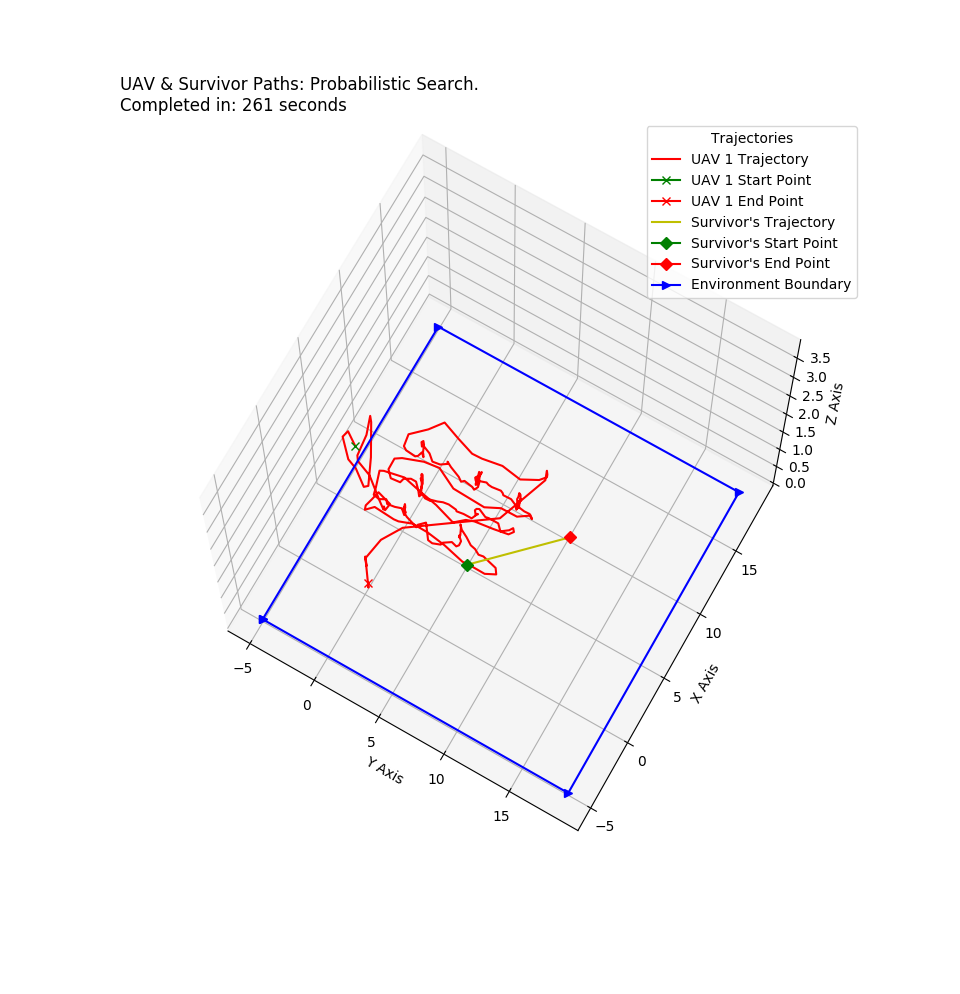}}}}%
    \qquad
    \subfloat{\fbox{{\includegraphics[scale=0.125]{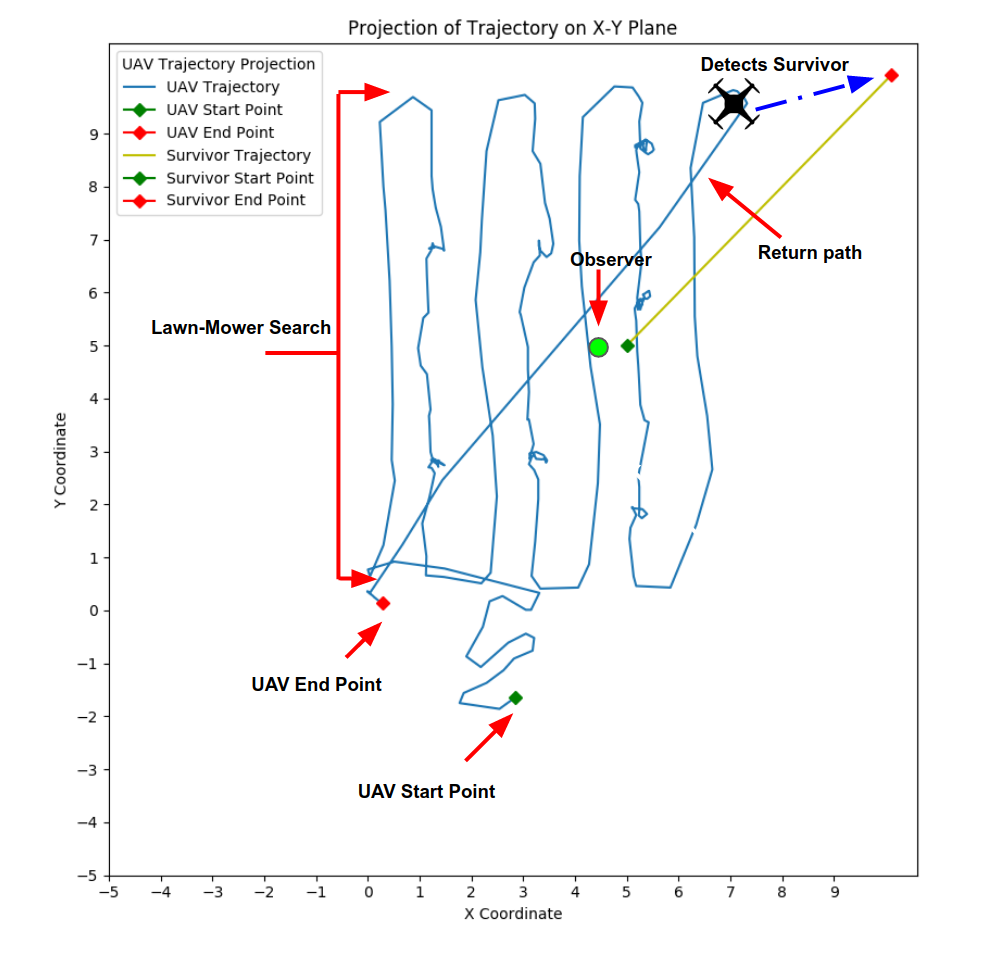}}}}%
    \caption{Trajectory and X-Y Projection of UAV and Survivor}%
    \label{fig:ProjectionPTLMXYZTrajectory}%
\end{figure}
\begin{figure}[h!]
    \centering
    \subfloat{\fbox{{\includegraphics[scale=0.15]{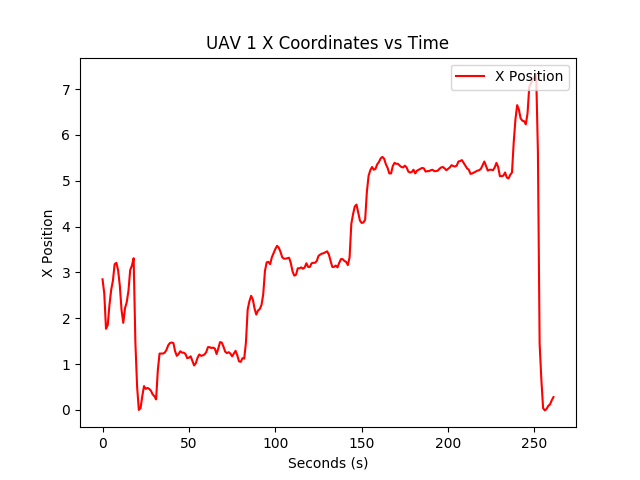} }}}%
    \qquad
    \subfloat{\fbox{{\includegraphics[scale=0.15]{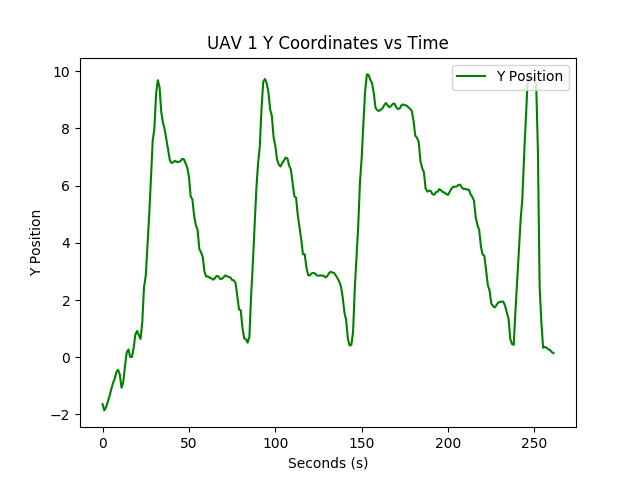} }}}%
    \qquad
    \subfloat{\fbox{{\includegraphics[scale=0.15]{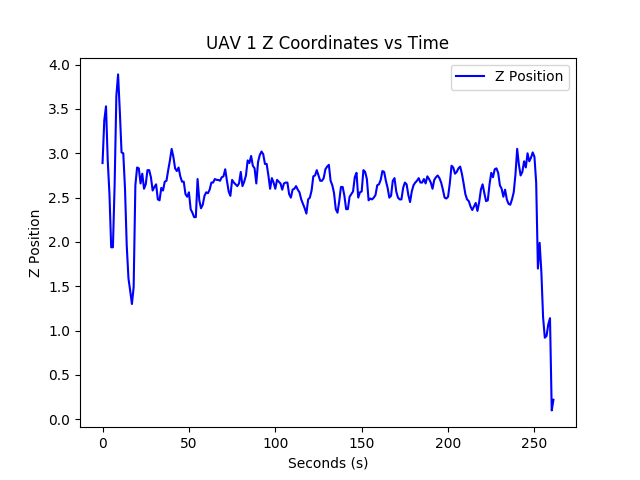} }}}%
    \caption{Variation in X, Y, Z Coordinates of Survivor \emph{V\textsubscript{s}} = 0.3 m/s}%
    \label{fig:PTLMXYZSurvivor}%
\end{figure}
\begin{figure}[h!]
    \centering
    \subfloat{\fbox{{\includegraphics[scale=0.15]{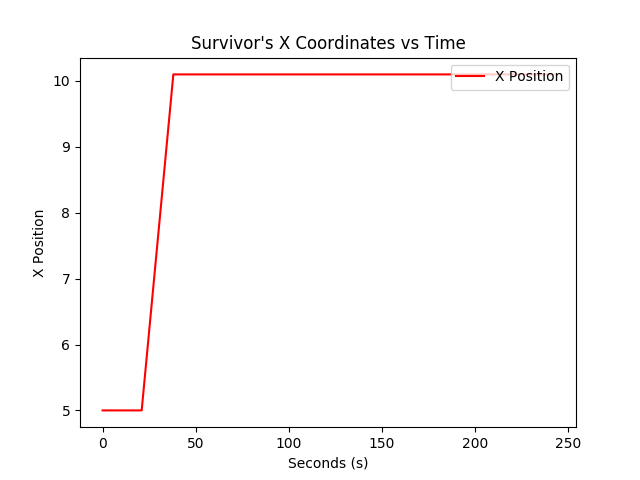} }}}%
    \qquad
    \subfloat{\fbox{{\includegraphics[scale=0.15]{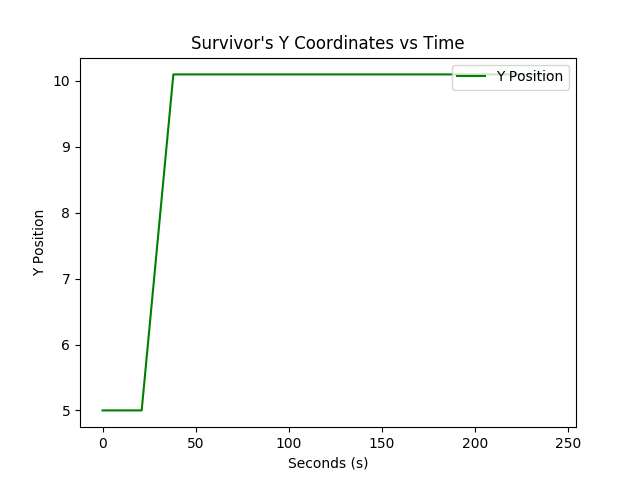} }}}%
    \qquad
    \subfloat{\fbox{{\includegraphics[scale=0.15]{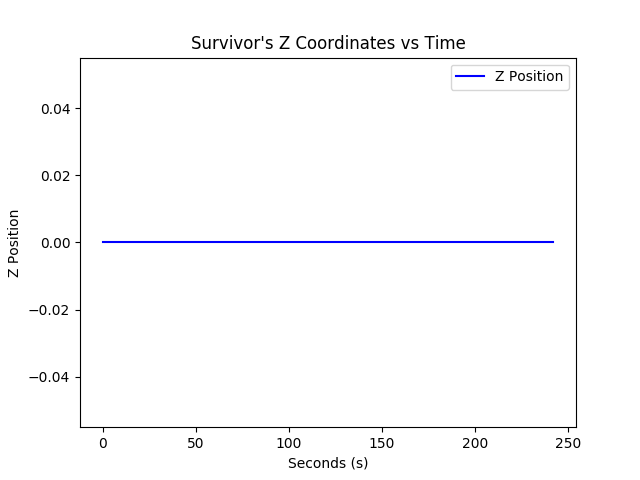} }}}%
    \caption{Variation in X, Y, Z Coordinates of UAV}%
    \label{fig:PTLMXYZUAV}%
\end{figure}

\noindent As seen form the Figure 1.19, 1.20 and 1.21, the lawn-mower search in physical testing takes 261 seconds to detect the survivor, in comparison to the 149 seconds taken by the weight-based exploration, an improvement of 75\textit{\%}.

\medskip

\noindent Inconsistent trajectory in Figures 1.17 and Figure 1.20 are observed due to strong external wind and also due to the erroneous measurements by the on-board GPS and barometer.

\medskip

\noindent The results can be improved by using a differential GPS for location accuracy and LiDAR for precise altitude measurements. Effects of external wind can be dealt with by incorporating a suitable controller.

\medskip

\noindent As apparent from the figures generated from the ROS simulations and the physical testing, our algorithm out-performs the traditional lawn-mower search. Furthermore, the results from varying environment sizes show that the weight-based algorithm is one that is agnostic of the environment size in which the UAV operates. 

\medskip

\noindent As inferred from the plots, the weight-based model outperforms the traditional lawn-mower search, given its utilization of the survivor's information to generated the weighted map of waypoints when the \texttt{observer\_node} was invoked. 

\medskip

\noindent The improvement in the search time of the weight-based model, both in simulation and physical testing, can be attributed to:

\begin{enumerate}
    \item Prioritization of waypoints using a probabilistic model of weight allocation, utilizing the survivor's information.
    \item Immediate relocation of the UAV to the survivor's last known position upon intimation by an on-ground observer.
\end{enumerate}

\noindent Given these two conditions, the weight-based model outperforms the traditional lawn-mower search. Furthermore, the model described in this chapter, under sub-section 1.2.1, is agnostic to the size of the environment in which the UAV operates; regardless of its dimensions, the waypoints within the environment can be assigned a weight and be prioritized for exploration. This allows us to carry out exploration in asymmetric environments as well.

\section{Conclusion and Future Work}

\noindent In this chapter, we present an implementation and improvement of a previously described "weight-based" exploration method. We implemented the model on ROS and an off-the-shelf UAV. 

\medskip

\noindent In comparison to the standard lawn-mower pattern of search the weight-based search, both in simulation and physical testing, demonstrates a significant improvement to the time taken to search for a survivor.

\medskip

\noindent The model described is agnostic to the number of agents and survivors. Our future work involves deploying this model on multiple agents to investigate large swaths of land for survivors, in collaboration with on-ground personnel.

\medskip

\noindent The physical tests detailed in the previous sections were restricted to controlled environments. Future tests will be conducted at the Indian Institute of Science's Challakere Campus, given the semblance to a realistic environment where such a solution would be deployed to aid first-responders.

\medskip

\noindent In this chapter, we've assumed a virtual survivor for the UAV to track; However, we are currently developing a novel computer-vision pipeline trained on images of humans from an overhead camera. This pipeline would be integrated into the current model and will be deployed to detect human survivors autonomously using an on-board camera.

\medskip

\noindent In the future, we aim to deploy this algorithm on a swarm of UAVs, which, along-with human counterparts would have the capability to investigate large swaths of flooded area, effectively speeding up the search for survivors.

\section*{Acknowledgements}

\noindent This work was partially supported by a Engineering and Physical Sciences Research Council - Global Challenges Research Fund (EPSRC-GCRF), UK, multi-institute grant (Grant Number: EP/P02839X/1).
%\begin{frontmatter}
%  \chapter*{References}
%  \markboth{References}{References}
%\end{frontmatter}

\end{document}